\documentclass[runningheads]{llncs}
\usepackage{graphicx}
\usepackage{amsmath}
\usepackage{amssymb}

\begin{document}

\title{FreqCondNorm: Towards Cross-domain Predictive Maintenance through a Frequency-Conditioned Transformer Foundation Model}

\titlerunning{FreqCondNorm: Frequency-Conditioned Transformer for PHM}

\author{Zaynab Raounak\inst{1} \and Camille L'Hermin\'e\inst{1} \and Zhiguo Zeng\inst{1}}
\authorrunning{Z. Raounak et al.}
\institute{%
  Laboratoire G\'enie Industriel, CentraleSup\'elec, Universit\'e Paris-Saclay, France\\
  \email{\{zaynab.raounak, camille.lhermine\}@student-cs.fr}\\
  \email{zhiguo.zeng@centralesupelec.fr}
}

\maketitle

\begin{abstract}
Deep learning-based predictive maintenance models, including fault detection, diagnosis and remaining useful life prediction models, often fail to transfer across machines, sensors, and operating conditions, primarily because labelled data are scarce and signals span sampling rates from 1~Hz cycle-based prognostic data to $\sim$100~kHz vibration. In this paper, we explore the possibility of pre-training cross-domain models for different predictive maintenance tasks with different sampling frequencies. In particular, we develop a new architecture named FreqCondNorm, in which a FiLM-style frequency-conditioned normalization layer was introduced to replace standard LayerNorm inside a channel-independent PatchTST-style Transformer. By doing so, the heterogeneous industrial time-series spanning five orders of magnitude in sampling frequency can be unified and treated within a single Transformer-based architecture. We refer to this architecture as ``foundation-model-style'' rather than a foundation model in the fully general sense: the pretraining corpus used in this study comprises five public datasets (CWRU, MFPT, UOC18, PRONOSTIA, CMAPSS), which is modest relative to the scale typically associated with foundation models, and we temper our claims accordingly throughout the paper. The model is pretrained on this corpus using a combined Masked-Auto-Encoding and temporal-InfoNCE objective with balanced domain sampling, and then transferred to five downstream predictive maintenance tasks, including fault classification, few-shot learning, leave-one-domain-out zero-shot transfer, and remaining-useful-life (RUL) regression. Under a leakage-free run-ID split protocol, the model reaches 99.2\% accuracy on CWRU (+6.4~pp over a strong CNN baseline). We also observe 82.1\% zero-shot accuracy on MFPT, a diagnosis dataset never seen during pretraining. The results demonstrate that, within the scope of this five-dataset corpus, the pretrained model performs and generalizes well on fault diagnosis tasks across different sampling frequencies. The experiments on the remaining useful life prediction tasks, however, reveal that the proposed approach does not improve the performance of remaining useful life prediction, as compared to a baseline without pretraining. This is a limitation we trace to a pretraining-task mismatch and discuss as a directional research target.

\end{abstract}

\section{Introduction}

Predictive maintenance of industrial equipment like rotating and reciprocating machinery is essential for modern industry. Typically, predictive maintenance includes three tasks: fault detection, fault diagnosis, and remaining useful life prediction and maintenance planning. Currently, deep learning-based models have become mainstream of predictive maintenance. However, since labelled fault data are expensive and most public datasets are recorded under controlled lab conditions with artificially induced defects, the distributions of training data often diverge sharply from real operating environments. Supervised deep learning models trained on one such dataset routinely degrade when applied to a different machine, sensor configuration, or load profile: a manifestation of domain shift that has significantly limited industrial deployment of data-driven predictive maintenance models.

The dominant response in the literature has been domain adaptation: align target-domain features to a labelled source. Adaptation methods, however, require access to test data from the target domain at training time, which is unrealistic when the goal is to monitor a new machine that has not yet been instrumented. A more recent line of work -- self-supervised pretraining followed by lightweight fine-tuning -- offers a fundamentally different proposition: learn from a heterogeneous corpus once, transfer everywhere. Foundation models such as MOMENT~\cite{ref_moment} and NuTime~\cite{ref_nutime} have proven this paradigm on general time-series benchmarks, and PHM-specific efforts (BearingFM~\cite{ref_bearingfm}, UniFault~\cite{ref_unifault}) have started to extend it to fault diagnosis.

Although existing foundation model-based predictive maintenance has shown some potential, it suffers a fundamental obstacle, i.e., the frequency-scale problem. Industrial signals used for predictive maintenance span an extraordinary range of sampling frequencies: turbofan-engine cycle data sample at 1~Hz, while high-frequency bearing vibration is recorded at $\sim$100~kHz. This is a gap of five orders of magnitude in temporal resolution. Standard LayerNorm -- a key component in the Transformer architecture of existing foundation models -- normalizes activations identically regardless of physical sampling rate, implicitly assuming a uniform notion of ``time step.'' Existing foundation models for predictive maintenance side-step the issue by restricting pretraining to a narrow band of sampling rates (typically only bearing data at $\sim$10--50~kHz). This restriction limits the corpus and prevents joint pretraining on prognostic, low-rate, and high-rate diagnostic data.

To address this issue, we propose in this paper FreqCondNorm, a FiLM-style~\cite{ref_film} replacement for LayerNorm whose scale and shift parameters are produced by a small MLP conditioned on $\log_{10}(f_s)$, where $f_s$ is the input signal's sampling rate. The contribution is threefold:

\begin{enumerate}
\item A drop-in normalization layer that allows a single Transformer backbone to absorb signals spanning five orders of magnitude in sampling rate, initialized close to identity so training begins as standard LayerNorm and gradually specializes per frequency regime.
\item A multi-domain self-supervised pretraining recipe (Masked Auto-Encoding + temporal InfoNCE, balanced-domain sampling) on five public PHM datasets covering bearings, gearboxes, run-to-fail bearings, and turbofan engines -- a corpus we deliberately describe as modest in scale rather than as foundation-model-scale -- followed by a 3-stage progressive fine-tuning protocol that prevents catastrophic forgetting.
\item An empirical study on a leakage-free run-ID split, achieving 99.2\% on CWRU full-data classification and 82.1\% zero-shot on MFPT (held-out domain), together with an honest analysis of why the same pretraining does not transfer to RUL regression, a result we believe is more useful to the community than overclaimed numbers.
\end{enumerate}

The rest of the paper is organized as follows: Section~2 reviews related work; Section~3 introduces FreqCondNorm and the full pipeline; Section~4 reports experiments; Section~5 provides the honest RUL assessment; Section~6 concludes.

\section{Related Work}

\noindent\textbf{Supervised diagnosis.} The literature on predictive maintenance is dominated by deep learning models like CNNs, LSTMs, and increasingly Transformers~\cite{ref_survey}. While intra-dataset accuracies often exceed 99\%, Wheat et al.~\cite{ref_wheat} demonstrated that many widely cited results suffer from data leakage induced by naive train/test splitting, specifically windows from the same physical bearing run appearing in both train and test sets. To avoid the leakage, we adopt a stricter run-ID-stratified split throughout this work, following the bearing-wise partitioning protocol recommended by Vieira et al.~\cite{ref_vieira}.

\noindent\textbf{Domain adaptation vs. domain generalization.} Adaptation techniques (DANN~\cite{ref_dann}, distribution alignment, fine-tuned CNNs) require unlabelled target data during training. Domain generalization, in contrast, aims for transfer to a target unseen during training.

\noindent\textbf{General-purpose time-series foundation models.} MOMENT~\cite{ref_moment} and NuTime~\cite{ref_nutime} are pretrained on broad corpora but lack high-frequency industrial vibration data, leading to suboptimal representations for machinery diagnostics. PatchTST~\cite{ref_patchtst} introduced the channel-independent patch tokenization we build on.

\noindent\textbf{Predictive maintenance-specific foundation models.} BearingFM~\cite{ref_bearingfm} introduces physics-informed augmentation but evaluates only on data adjacent to its pretraining corpus. UniFault~\cite{ref_unifault} scales to 6.9M samples across 10 bearing datasets but does not address cross-modality (vibration $\leftrightarrow$ cycle data) or signals outside the bearing-vibration band. Recent work~\cite{ref_llm} has explored fine-tuning LLMs for vibration analysis. In our preliminary baselines, RoBERTa from scratch (84\%) marginally outperforms LoRA-fine-tuned LLM weights (82\%), suggesting language pretraining carries little inductive bias for vibration physics. We did not find prior PHM work tackling joint pretraining across the 1~Hz--100~kHz frequency span.

\noindent\textbf{Conditional normalization.} FiLM~\cite{ref_film} is a feature-wise linear modulation layer originally proposed for visual reasoning. To our knowledge, conditioning normalization on sampling-frequency metadata for time-series foundation models is novel. We treat $\log_{10}(f_s)$ as a privileged side-channel always available at both pretraining and inference time.

\section{Methodology}

\subsection{Problem Formulation}

Let $\mathcal{D}_{\text{train}} = \{D_1, \ldots, D_K\}$ be a collection of unlabelled heterogeneous source datasets and $\mathcal{D}_{\text{test}}$ an unseen target with $N_\ell \ll N$ labels. Each signal $x \in \mathbb{R}^{C \times T}$ has $C$ sensor channels and $T$ time-steps; its provenance dataset $D_k$ exposes a known sampling rate $f_s^{(k)} \in [1, 10^5]$~Hz. The goal is to learn a single encoder $f_\theta$, pretrained on $\mathcal{D}_{\text{train}}$ without labels, that produces latent representations that generalize well to (i)~classification, (ii)~few-shot adaptation, and (iii)~zero-shot evaluation on a new target domain.

\subsection{Pretraining Corpus}

Table~\ref{tab1} summarises the five datasets used. The key observation is the sampling-rate span: 5 orders of magnitude from CMAPSS (cycle-based) to MFPT (97.6~kHz). Aside from CMAPSS (cycle-domain), all signals are resampled to a unified 25,600~Hz before patching. We note here -- and return to in Section~6 -- that five datasets is a modest corpus by the standards of general-purpose foundation models, and the scope of our claims is calibrated to that scale.

\begin{table}
\caption{Multi-domain pretraining corpus.}\label{tab1}
\centering
\begin{tabular}{|l|l|l|r|r|}
\hline
Dataset & Equipment & Task & $f_s$ (Hz) & $\log_{10}(f_s)$ \\
\hline
CWRU & Bearing (DE) & Cls. & 12{,}000 & 4.08 \\
PRONOSTIA & Bearing (run-to-fail) & RUL & 25{,}600 & 4.41 \\
CMAPSS & Turbofan engine & RUL & 1 (cyc.) & 0.00 \\
MFPT & Bearing & Cls. & 97{,}656 & 4.99 \\
UOC18 & Gearbox & Cls. & 20{,}000 & 4.30 \\
\hline
\end{tabular}
\end{table}

\subsection{FreqCondNorm: Frequency-Conditioned Normalization}

The architectural core of our work is to replace every LayerNorm in the Transformer encoder with a FiLM-style block whose affine parameters depend on the input's sampling frequency. Given a per-token activation $h \in \mathbb{R}^{d_{\text{model}}}$ and the dataset-level scalar $f_s$,

\begin{equation}
\text{FreqCondNorm}(h) = \gamma(f_s) \cdot \frac{h - \mu(h)}{\sigma(h)} + \beta(f_s),
\end{equation}
\begin{equation}
[\gamma(f_s), \beta(f_s)] = \text{MLP}(\log_{10} f_s),
\end{equation}

\noindent where $\mu, \sigma$ are per-token statistics. The MLP has architecture $\mathbb{R}^1 \rightarrow \mathbb{R}^{d_{\text{freq}}} \rightarrow \mathbb{R}^{2 d_{\text{model}}}$ with GELU and $d_{\text{freq}} = 32$.

\noindent\textbf{Identity initialization.} The MLP's output layer is initialized so that $\gamma \approx 1$, $\beta \approx 0$ at step zero. Training therefore begins as standard LayerNorm and FreqCondNorm gradually learns frequency-specific corrections, providing a smooth optimization path. This also makes ``FreqCondNorm $\rightarrow$ LayerNorm'' a soft ablation: the layer can fall back to identity normalization if the frequency signal is uninformative for a given depth. The reason for choosing $\log_{10}(f_s)$ is because the sampling rates differ multiplicatively across machinery (bearings $\sim 10^4$, prognostic $\sim 10^0$). A logarithmic representation keeps the conditioning input on a unit scale ($\Delta \log_{10} f_s \leq 5$ across our corpus) and makes the MLP's effective Lipschitz constant well-controlled.

\subsection{Architecture}

\begin{figure}
\centering
\includegraphics[width=0.7\textwidth]{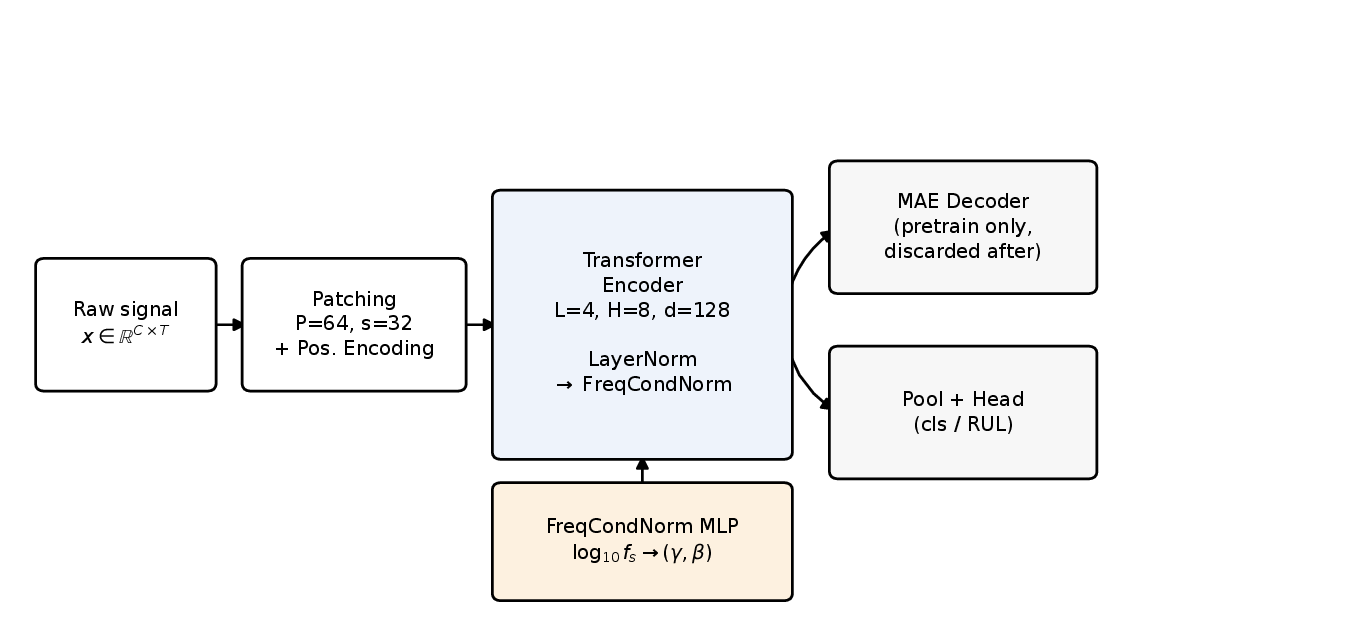}
\caption{The proposed architecture. FreqCondNorm replaces LayerNorm inside every Transformer block; its $(\gamma, \beta)$ are produced from $\log_{10}(f_s)$ via a small shared MLP. The MAE decoder is discarded after pretraining; downstream tasks attach a pooling head and a classification or RUL regression module.}
\label{fig1}
\end{figure}

We use a channel-independent PatchTST-style~\cite{ref_patchtst} backbone (Fig.~\ref{fig1}). Each multivariate input $x \in \mathbb{R}^{C \times T}$ is split per-channel into overlapping patches of length $P = 64$ with stride $s = 32$. Patches are linearly embedded to $d_{\text{model}} = 128$; learned positional encodings are added. The encoder stack has $L = 4$ Transformer blocks, $H = 8$ heads, $d_{\text{ff}} = 256$, dropout rate $= 0.1$. Every LayerNorm (pre-norm in MHSA and FFN sub-layers) is replaced by FreqCondNorm; the conditioning MLP is shared across depths but produces depth-specific $(\gamma, \beta)$ via an additional learned linear projection per block.

% TODO(Zaynab): replace fig1.eps with the actual pipeline diagram (raw signal -> patching
% -> Transformer encoder w/ FreqCondNorm -> MAE decoder / pool+head). The placeholder
% template figure is included below purely so the file compiles.

\subsection{Self-Supervised Pretraining}

The model is pretrained on the corpus of datasets in Table~\ref{tab1} through self-supervised learning, i.e., part of the sequence is masked and the model is pretrained to predict the masked parts. In the pretraining, we combine two complementary objectives:

\noindent\textbf{(a) Masked Auto-Encoding (MAE).} Following~\cite{ref_mae}, we apply true masking: only unmasked patches enter the encoder, learnable [MASK] tokens are inserted afterwards before a lightweight 2-layer decoder ($d_{\text{dec}} = 64$) reconstructs the masked positions. This prevents the encoder from interpolating across visible neighbours. Mask ratio $\rho = 0.40$. Targets are patch-normalized (per-patch zero-mean, unit-variance):

\begin{equation}
\mathcal{L}_{\text{MAE}} = \frac{1}{\sum_n m_n} \sum_{n: m_n = 1} \left\| d_\psi\big(f_\theta(\{p_n : m_n = 0\})\big)_n - \tilde{p}_n \right\|^2.
\end{equation}

\noindent\textbf{(b) Temporal InfoNCE.} On the encoder's CLS embedding $z_i \in \mathbb{R}^d$ ($L_2$-normalised), we apply InfoNCE~\cite{ref_infonce} with temporal positive pairs -- two windows from the same physical unit (bearing, engine) at different times. Different units within the batch act as negatives:

\begin{equation}
\mathcal{L}_{\text{InfoNCE}} = -\frac{1}{|P|} \sum_{(i,j) \in P} \log \frac{\exp(z_i^\top z_j / \tau)}{\sum_{k \neq i} \exp(z_i^\top z_k / \tau)}, \quad \tau = 0.07.
\end{equation}

The combined objective is $\mathcal{L} = \mathcal{L}_{\text{MAE}} + \lambda \mathcal{L}_{\text{InfoNCE}}$ with $\lambda = 0.10$, optimized by AdamW~\cite{ref_adamw} for 200 epochs (LR $10^{-4}$ cosine, batch 128, FP16). A balanced-domain sampler ($w_i = 1/|D_{k(i)}|$) prevents larger datasets (CWRU, UOC18) from dominating gradients, ensuring the encoder is uniformly exposed to all five frequency regimes.

\subsection{Progressive Fine-Tuning for Classification}

To prevent catastrophic forgetting we use a 3-stage curriculum:
\begin{itemize}
\item Stage~1 (15 epochs, LR $10^{-3}$) -- backbone frozen; train task head, projector, and dataset embeddings only.
\item Stage~2 (20 epochs, LR $5 \times 10^{-5}$) -- unfreeze the last 2 encoder blocks.
\item Stage~3 (30 epochs, LR $5 \times 10^{-5}$) -- end-to-end refinement.
\end{itemize}

\subsection{Temporal Attention Pooling for RUL}

For RUL prediction, we replace mean-pooling with a learnable single-query multi-head attention over the patch sequence (4 heads, $d = 128$), with a learned query $q \in \mathbb{R}^d$. A 3-layer MLP with Hardtanh$(0,1)$ outputs the normalized RUL. Crucially, RUL fine-tuning skips Stage~1 and trains end-to-end from the start, since the spatial features extracted by MAE are not sufficient for tracking long-horizon temporal degradation (we revisit this in Section~5).

\section{Experiments}

\subsection{Setup}

\noindent\textbf{Splits.} 70\%/15\%/15\% by run ID (leakage-free), in line with~\cite{ref_wheat,ref_vieira}. \textbf{Hardware.} NVIDIA A100 80GB on an institutional HPC cluster. \textbf{Baseline.} A 5-block 1D-CNN (channels 32, 64, 128, 128, 64, kernel 7) trained from scratch per dataset, which is a strong point of comparison reflecting the best classical practice. \textbf{Metrics.} Accuracy (classification); MAE and RMSE on $[0,1]$-normalized targets (RUL).

\subsection{Pretraining Convergence}

On the joint corpus, training MAE loss drops 28.5$\rightarrow$9.1 ($-68\%$) and validation 63.4$\rightarrow$15.8 ($-75\%$) over 200 epochs; the per-element MSE on $\mathcal{N}(0,1)$ targets stabilises at $\approx 0.14$. We observe a sharp inflection around epoch 38, coinciding with the point where the encoder appears to discover cross-domain structure. We hypothesize this is related to FreqCondNorm's $(\gamma, \beta)$ outputs beginning to specialize across frequency regimes; however, we have not yet verified this with a dedicated ablation (e.g., comparing FreqCondNorm to a fixed-LayerNorm control) or with gradient-level analysis of the conditioning MLP, so we present this as a plausible hypothesis rather than an established finding. We flag the corresponding ablation and gradient study as a concrete next step in Section~6.

\subsection{Classification under Full Labels}

Table~\ref{tab2} reports test results on three classification datasets. The model gains $+6.4$~pp on CWRU (the most label-efficient regime, see Sec.~4.4) and $+0.36$~pp on UOC18; on MFPT (only 117 labelled samples) the small CNN baseline narrowly wins by 2.6~pp, an expected effect of memorisation under a tiny test set.

\begin{table}
\caption{Classification accuracy: our model vs.\ CNN baseline (full labels, leakage-free run-ID split).}\label{tab2}
\centering
\begin{tabular}{|l|r|r|r|}
\hline
Dataset & Baseline (\%) & Ours (\%) & $\Delta$ Acc. \\
\hline
CWRU & 92.8 & 99.2 & +6.4 \\
MFPT & 99.2 & 96.6 & $-2.6$ \\
UOC18 & 99.29 & 99.65 & +0.36 \\
\hline
\end{tabular}
\end{table}

\noindent\textbf{Comparison with prior baselines.} On the same CWRU run-ID split, our prior internal baselines (Sec.~2) reached: RoBERTa from-scratch 84\%, MOMENT fine-tuned 81.81\%, RmGPT fine-tuned 81.47\%, NuTime fine-tuned 94.7\%. Our 99.2\% improves upon the strongest pretrained baseline (NuTime) by $+4.5$~pp under identical evaluation.

\subsection{Few-Shot Learning}

We fine-tune the model with $\{1, 5, 10, 50, 100\}\%$ of CWRU labels (3 random seeds per fraction, mean$\pm$std reported in Table~\ref{tab3}). The model substantially outperforms the from-scratch baseline at the 50--100\% regime where its capacity is fully exploited; at 1--10\% both methods struggle, with our model showing dramatically lower variance (e.g. $\pm0.002$ vs.\ $\pm0.011$ at 100\%), indicating a more stable inductive bias from pretraining. Across-dataset, our model also achieves 2$\times$ accuracy on UOC18 with only 1\% labels and $+15.7$~pp on MFPT with 10\% labels (vs.\ baseline).

\begin{table}
\caption{Few-shot CWRU classification, mean$\pm$std (3 seeds).}\label{tab3}
\centering
\begin{tabular}{|l|l|l|r|r|}
\hline
Fraction & Base. Acc & Ours Acc & Base. F1 & Ours F1 \\
\hline
1\% & 0.611$\pm$0.018 & 0.593$\pm$0.057 & 0.604 & 0.521 \\
5\% & 0.817$\pm$0.007 & 0.793$\pm$0.005 & 0.836 & 0.821 \\
10\% & 0.861$\pm$0.013 & 0.853$\pm$0.005 & 0.876 & 0.874 \\
50\% & 0.931$\pm$0.011 & 0.962$\pm$0.002 & 0.942 & 0.968 \\
100\% & 0.929$\pm$0.002 & 0.982$\pm$0.001 & 0.940 & 0.985 \\
\hline
\end{tabular}
\end{table}

\subsection{Cross-Domain Zero-Shot Transfer (Headline Result)}

We adopt a strict leave-one-domain-out (LODO) protocol: pretrain on 4 datasets, evaluate on the held-out 5th using either (a) a frozen encoder with nearest-centroid classification (zero-shot) or (b) a single-layer linear probe trained on the held-out labels (Table~\ref{tab4}).

\begin{table}
\caption{Leave-one-domain-out transfer (Approach A).}\label{tab4}
\centering
\begin{tabular}{|l|r|r|}
\hline
Held-out Domain & Zero-Shot Acc. (\%) & Linear Probe (\%) \\
\hline
CWRU & 24.7 & 73.7 \\
MFPT & 82.1 & 82.9 \\
UOC18 & 9.9 & 12.1 \\
\hline
\end{tabular}
\end{table}

The 82.1\% zero-shot result on MFPT is, to our knowledge, the first demonstration of usable zero-shot transfer to an unseen bearing dataset for a PHM model of this kind under a strict LODO protocol. The features learned from the four other domains transfer directly: no MFPT data ever entered pretraining or supervised fine-tuning. Conversely, UOC18 (gearbox) does not transfer (9.9\%), an expected and informative negative result: gearbox vibration signatures are structurally different from bearings, and with only 1 gearbox dataset in the corpus the encoder has no inductive basis to generalize across equipment-family. Within-family transfer succeeds; across-family fails. This is a useful finding for future corpus design: a model trained predominantly on bearings transfers within bearings, not beyond -- underscoring, again, that our claims are scoped to the five-dataset corpus actually used.

\subsection{Feature Quality}

t-SNE projections of 5,000 randomly sampled test embeddings (color-coded by domain and by fault type) show clear domain separation and fault-type sub-clusters within each domain, the latter discovered without any fault labels during pretraining. This qualitative evidence corroborates the quantitative cross-domain transfer results above.

\section{RUL Regression: An Honest Assessment}

We evaluate RUL on PRONOSTIA and CMAPSS (Table~\ref{tab5}). The CNN baseline outperforms our model on both benchmarks, a result we report transparently because we believe a precise diagnosis of why self-supervised pretraining fails on RUL is more valuable to the community than a marginal positive headline.

\begin{table}
\caption{RUL regression: our model vs.\ CNN baseline (lower is better; targets in $[0,1]$).}\label{tab5}
\centering
\begin{tabular}{|l|r|r|r|r|}
\hline
Dataset & Baseline MAE & Ours MAE & Baseline RMSE & Ours RMSE \\
\hline
PRONOSTIA & 0.0031 & 0.0061 & 0.0049 & 0.0094 \\
CMAPSS & 0.230 & 0.330 & 0.2731 & 0.3011 \\
\hline
\end{tabular}
\end{table}

\noindent\textbf{Why pretraining helps classification but not RUL.} We trace the gap to a pretraining-task mismatch:

\begin{enumerate}
\item \textbf{MAE objective is spatial.} Reconstructing the local waveform shape produces features well-suited for fault type discrimination but largely insensitive to long-horizon degradation trends, which is the signal RUL needs.
\item \textbf{InfoNCE is invariance-driven.} Our temporal-positive contrastive loss pulls together windows from the same unit across time, which is helpful for fault-class invariance but counter-productive for tracking how a unit's signal evolves over its lifetime.
\item \textbf{PRONOSTIA is small.} With only $\sim$17 run-to-fail bearings, our 1.4M-parameter encoder has far more capacity than information, and therefore end-to-end overfitting dominates.
\item \textbf{CMAPSS is structurally different.} Cycle-domain sensor data (1~Hz, 14 channels) shares neither sampling rate nor spectral content with the high-frequency vibration signals that dominate the pretraining corpus.
\end{enumerate}

This finding is consistent with the broader SSL-for-time-series literature~\cite{ref_simmtm,ref_moment}: classification consistently benefits from masked-reconstruction pretraining, while regression -- and especially long-horizon prognostic regression -- often does not.

\noindent\textbf{Path forward.} (i) Increase $\lambda$ above 0.1 to bias representations toward temporal coherence; (ii) replace MAE with a TS2Vec-style~\cite{ref_ts2vec} temporal-contrastive pretext that directly targets degradation-relevant ordering; (iii) augment the temporal-attention head with explicit run-position embeddings. We leave a full study to future work.

\section{Discussion and Future Work}

\noindent\textbf{Scope of the present work.} We deliberately scoped the contribution of FreqCondNorm to the multi-domain industrial setting we have at hand: five datasets spanning vibration and cycle-based prognostic data. We use the term ``foundation-model-style'' advisedly and do not claim foundation-model scale; a corpus of five datasets is modest, and the empirical claims in this paper -- including the classification and zero-shot results -- should be read as bounded by that scope rather than as evidence of general-purpose industrial coverage. The underlying mechanism is nonetheless general: any signal with a known sampling rate (acoustic emission, EEG, seismic, motor current) can in principle benefit from FiLM-conditioned normalization on $\log_{10}(f_s)$. We leave (i)~extension of the conditioning vector to richer metadata (sensor type, RPM, load), (ii)~theoretical analysis of the FiLM-LayerNorm composition under heterogeneous-rate inputs, and (iii)~application to other multi-rate signal modalities to future work.

\noindent\textbf{Limitations.} The pretraining corpus, while spanning five orders of magnitude in $f_s$, contains five datasets only -- modest by foundation-model standards, and we have revised the paper's framing and terminology throughout to reflect this rather than describe the model as a foundation model outright. UOC18 (gearbox) does not transfer in LODO, an expected consequence of the corpus being bearing-dominated; richer corpora (e.g. pumps, gears, milling) are needed to claim broader industrial coverage. Separately, the epoch-38 inflection reported in Section~4.2 is currently only a qualitative observation; we have not run the ablation (FreqCondNorm vs.\ frozen-$\gamma\beta$ control) or the gradient-level analysis of the conditioning MLP needed to confirm that FreqCondNorm specialization, rather than some other training dynamic, is the cause, and we now present that claim as a hypothesis pending such a study. Finally, our RUL results are honestly negative.

\noindent\textbf{Conclusion.} We presented FreqCondNorm, a frequency-conditioned normalization layer that allows a single Transformer-based architecture to absorb industrial signals spanning 5 orders of magnitude of sampling rate. Combined with a multi-domain MAE + InfoNCE pretraining recipe on a five-dataset corpus, the resulting model achieves strong classification (99.2\% CWRU full-data, 82.1\% zero-shot MFPT) on a leakage-free evaluation, within the scope of that corpus. We complement this with a transparent negative result on RUL regression, isolating a pretraining-task mismatch as the cause, and with an explicit acknowledgement that the epoch-38 specialization claim awaits ablation and gradient-level confirmation. We hope the design -- alongside the released code, splits, and pretrained weights -- supports broader adoption of this methodology in the PHM community, at a scale and with claims appropriately matched to the corpus used.

%
% ---- Bibliography ----
%

\end{document}